\documentclass{article}

 \usepackage[preprint]{neurips_2026}

\usepackage[utf8]{inputenc} 
\usepackage[T1]{fontenc}    
\usepackage{hyperref}       
\usepackage{url}            
\usepackage{booktabs}       
\usepackage{amsfonts}       
\usepackage{nicefrac}       
\usepackage{microtype}      
\usepackage[table]{xcolor}  
\usepackage{graphicx}
\usepackage{subcaption}
\usepackage{booktabs}
\usepackage{multirow}
\usepackage{comment}

\usepackage{amsmath}
\usepackage{amssymb}
\usepackage{mathtools}
\usepackage{amsthm}
\usepackage[capitalize,noabbrev]{cleveref}

\usepackage{algorithm}
\usepackage{algorithmic}
\usepackage{enumitem}

\usepackage{pifont}
\newcommand{\cmark}{\ding{51}}%
\newcommand{\xmark}{\ding{55}}%
\newcommand{\pmark}{{\raise.17ex\hbox{$\scriptstyle\sim$}}}

\title{Memory-Efficient Integer-Only Training\\ of Deep Neural Networks}

\author{
  Hazem Hesham Yousef Shalby\thanks{Corresponding author: \texttt{hazemhesham.shalby@polimi.it}}
  \quad Fabrizio Pittorino \quad Manuel Roveri \\
  Dipartimento di Elettronica, Informazione e Bioingegneria (DEIB)\\
  Politecnico di Milano, Milan, Italy
}

\begin{document}

\maketitle

\begin{abstract}
Training neural networks in reduced precision can yield significant efficiency gains, yet all existing methods that scale to standard computer vision benchmarks retain floating-point in at least one stage, typically the weight update or optimizer state. We present a training algorithm that scales to modern deep neural networks, in which every tensor (weights, gradients, and activations) is stored as INT8, and every operation (forward pass, backward pass, weight update, optimizer update) is performed in integer arithmetic. Specifically, tensors are stored as signed integer mantissas paired with explicit power-of-two exponents, on which integer-only operations are performed using five custom composable building blocks (products, sums, divide-by-constant, nonlinear functions, and weight updates). This allows complete implementation of trainable integer layers, including convolutions, batch and layer normalization, softmax, and GELU. A per-group velocity mechanism with independent exponent tracking preserves gradient precision across neurons and channels with diverse magnitudes. As every tensor fits in INT8, training memory is roughly $4\times$ smaller than the FP32 baseline. We validate our training algorithm on CNNs (ResNet-18/34/50/101, MobileNetV2) and Vision Transformers (ViT-Ti / B), demonstrating  for the first time that training neural networks with INT8 tensors and integer arithmetic can achieve competitive accuracy on CIFAR-10, CIFAR-100 and ImageNet at a reduced memory budget.
\end{abstract}

\section{Introduction}
\label{sec:intro}

The cost of training deep neural networks continues to grow with model size and dataset scale.
A natural way to reduce this cost is to lower numerical precision, replacing 32-bit floating-point arithmetic with narrower integer operations that are cheaper in energy, area, and latency.

Mixed-precision training~\cite{micikevicius2018mixedprecisiontraining} showed that FP16 suffices for forward and backward propagation with minimal accuracy loss, and later methods pushed the datapath further to FP8~\cite{micikevicius2022fp8formatsdeeplearning} and INT8~\cite{zhu2019unifiedint8trainingconvolutional,zhao2021distributionadaptiveint8quantization}.
However, these approaches still rely on floating-point arithmetic in at least one critical component, such as the weight update, optimizer state, or nonlinear functions.

This residual dependence on floating-point arithmetic remains a practical bottleneck for two complementary reasons. On GPUs, although floating-point computation is efficient, the persistent training state, including weights, gradients, and optimizer moments, still requires 12 bytes per parameter. As a result, model and batch size are often constrained by FP32 optimizer state rather than by the low-precision compute path itself. On integer-only platforms, including microcontrollers, fixed-function accelerators, and in-memory compute substrates, floating-point support may be absent entirely. In this setting, eliminating all floating-point operations is a fundamental prerequisite for on-device learning. Existing approaches that move toward fully integer training \cite{Wang_2022,hu2025tintintinylearningtiny} either do not include standard optimization mechanisms such as momentum and learning-rate schedules, or remain restricted to small models and tasks.

We address these challenges through a unified integer-only training framework.
Our key observation is that any tensor can be represented as a signed integer mantissa together with an explicit power-of-two exponent, so arithmetic remains in the integer domain while scale is tracked exactly through exponent addition.
From this representation, we derive five reusable primitives: products, sums, divide-by-constant, nonlinear functions, and weight updates.
These primitives compose into complete layers with integer-only forward and backward computation.
To preserve gradient precision when magnitudes vary substantially within a weight tensor, we further introduce a per-group velocity mechanism that maintains separate exponent tracking for different parameter groups, such as output channels, and enables stable integer-only SGD with momentum.

Together, these ideas define, to our knowledge, the first fully integer-only training framework spanning forward propagation, backward propagation, and parameter updates that scales to modern deep neural networks and standard computer vision benchmarks.
The framework is general enough to realize standard CNN and ViT layers, including integer-only softmax, GELU, batch and layer normalization.

\section{Related Work}
\label{sec:related}

We organize prior work by how far each approach eliminates floating-point from training; \Cref{tab:method_comparison} summarizes the comparison and \Cref{tab:storage_formats} lists the per-tensor storage format of each method.

\paragraph{Low-precision training.}
Mixed-precision training~\cite{micikevicius2018mixedprecisiontraining} showed that FP16 suffices for forward/backward passes if FP32 master weights are maintained.
Subsequent methods reduced the datapath to FP8~\cite{micikevicius2022fp8formatsdeeplearning}.
Quantization-aware training methods (PACT~\cite{choi2018pactparameterizedclippingactivation}, LSQ~\cite{esser2020learnedstepsizequantization}, DoReFa-Net~\cite{zhou2018dorefanettraininglowbitwidth}, F8Net~\cite{jin2022f8netfixedpoint8bitmultiplication}, DQT~\cite{shalby2026dqt}) quantize the forward pass while keeping FP32 master weights and a floating-point backward; \citet{jacob2018quantizationtrainingneuralnetworks} introduced an integer-arithmetic-only inference framework adopted in mobile deployment.
All methods retain floating-point for the weight update.

\paragraph{Integer forward and backward passes.}
\citet{banner2018scalablemethods8bit} were the first to quantize weights, activations, and a substantial portion of the gradient stream to 8-bit on ImageNet, though the final weight-gradient accumulation remains in higher precision.
\citet{zhu2019unifiedint8trainingconvolutional} introduced gradient clipping strategies for full INT8 training on ImageNet-scale CNNs; \citet{zhao2021distributionadaptiveint8quantization} added per-channel scaling to close the accuracy gap.
Fixed-point variants adapt bit-widths dynamically~\cite{9157439}, while AMPA~\cite{pmlr-v235-ding24b} pushes inference below INT8 via mixed-precision allocation.
All leave the optimizer in FP32.

\paragraph{Quasi-integer training.}
SWALP~\cite{yang2019swalpstochasticweight} trains in 8-bit block floating-point, quantizing weights, activations, gradients, and the optimizer state via stochastic rounding; a higher-precision stochastic-weight-average running mean remains in float.
WAGE~\cite{wu2018traininginferencewithintegers} was among the first to discretize weights, activations, gradients, and errors to low-bit integers simultaneously, but replace batch normalization layers with constant scaling to avoid floating-point operations.
NITI~\cite{Wang_2022} performs all three stages (forward, backward, weight update) in integer arithmetic with a single per-layer block exponent, but this shared exponent forces small-gradient channels to lose their effective integer range once the scale adapts to large-gradient channels; the resulting sub-less-significant-bit (sub-LSB) vanishing is the failure mode that our per-group velocity (\Cref{sec:pergroup}) addresses. NITI also lacks momentum and LR scheduling, and in practice scales only to AlexNet.
\citet{ghaffari2023integerarithmeticdeeplearning} propose INT8 mantissas with an integer weight update, but their reference implementation relies on additional operations to switch back-and-forth from the floating point representation to the integer one, and falls back to float softmax in the ViT attention block.
Tin-Tin~\cite{hu2025tintintinylearningtiny} targets microcontrollers but has been applied only to tiny sensor tasks.

\paragraph{Integer-only nonlinear functions.}
On the inference side, I-BERT~\cite{pmlr-v139-kim21d} demonstrated that polynomial approximations can replace all nonlinear functions in BERT with integer-only arithmetic, and I-ViT~\cite{li2023ivit} extended this to Vision Transformers with integer-only Softmax, GELU, and LayerNorm.
We follow a different approach, by encoding all non-linear functions in look-up tables (LUTs), without significant memory overhead (see \Cref{sec:op_nonlinear}).


\begin{table}[t]
\centering
\caption{Comparison of low-precision training methods.
\cmark\ = fully integer / supported;
\xmark\ = floating-point / not supported;
\pmark\ = partially integer (some float remains).
\textbf{CNN @ INet} means ResNet-18--scale or CNNs trained/evaluated on ImageNet; \textbf{ViT} means training or fine-tuning of a Vision Transformer.}
\label{tab:method_comparison}
\small
\setlength{\tabcolsep}{4pt}
\begin{tabular}{p{1.9cm}lcccccc}
\toprule
\textbf{Category}
  & \textbf{Method}
  & \rotatebox{70}{\textbf{Fwd/Bwd int}}
  & \rotatebox{70}{\textbf{Wt update int}}
  & \rotatebox{70}{\textbf{Wt storage int}}
  & \rotatebox{70}{\textbf{Momentum + LR}}
  & \rotatebox{70}{\textbf{CNN @ INet}}
  & \rotatebox{70}{\textbf{ViT}} \\
\midrule
\multirow{4}{=}{\itshape Low-precision training}
  & Mixed-Prec.~\cite{micikevicius2018mixedprecisiontraining}
  & \xmark & \xmark & \xmark & \cmark & \cmark & \cmark \\
  & FP8~\cite{micikevicius2022fp8formatsdeeplearning}
  & \xmark & \xmark & \xmark & \cmark & \cmark & \cmark \\
  & F8Net~\cite{jin2022f8netfixedpoint8bitmultiplication}
  & \xmark & \xmark & \xmark & \cmark & \cmark & \xmark \\
  & DQT~\cite{shalby2026dqt}
  & \xmark & \xmark & \xmark & \cmark & \cmark & \xmark \\
\midrule
\multirow{5}{=}{\itshape Integer fwd/bwd passes}
  & \citet{banner2018scalablemethods8bit}
  & \cmark & \xmark & \xmark & \cmark & \cmark & \xmark \\
  & \citet{zhu2019unifiedint8trainingconvolutional}
  & \cmark & \xmark & \xmark & \cmark & \cmark & \xmark \\
  & \citet{zhao2021distributionadaptiveint8quantization}
  & \cmark & \xmark & \xmark & \cmark & \cmark & \xmark \\
  & \citet{9157439}
  & \cmark & \xmark & \xmark & \cmark & \cmark & \xmark \\
  & AMPA~\cite{pmlr-v235-ding24b}
  & \cmark & \xmark & \xmark & \cmark & \cmark & \xmark \\
\midrule
\multirow{5}{=}{\itshape Quasi-integer training}
  & WAGE~\cite{wu2018traininginferencewithintegers}
  & \cmark & \pmark & \pmark & \pmark & \xmark & \xmark \\
  & SWALP~\cite{yang2019swalpstochasticweight}
  & \pmark & \cmark & \pmark & \cmark & \cmark & \xmark \\
  & NITI~\cite{Wang_2022}
  & \cmark & \cmark & \cmark & \xmark & \xmark & \xmark \\
  & \citet{ghaffari2023integerarithmeticdeeplearning}
  & \cmark & \pmark & \xmark & \cmark & \cmark & \pmark \\
  & Tin-Tin~\cite{hu2025tintintinylearningtiny}
  & \cmark & \cmark & \cmark & \xmark & \xmark & \xmark \\
\midrule
\itshape \textbf{Integer-only}
  & \textbf{Ours}
  & \cmark & \cmark & \cmark & \cmark & \cmark & \cmark \\
\bottomrule
\end{tabular}
\end{table}

\begin{table}[t]
\centering
\caption{Per-tensor storage format and arithmetic precision.}
\label{tab:storage_formats}
\small
\setlength{\tabcolsep}{4pt}
\begin{tabular}{lcccc}
\toprule
\textbf{Method}
  & \textbf{Weights} 
  & \textbf{Gradient}
  & \textbf{Velocity}
  & \textbf{Activations}\\
\midrule
FP32                          & FP32 & FP32 & FP32  & FP32 \\
FP16~\cite{micikevicius2018mixedprecisiontraining}
                              & FP32 & FP32 & FP32  & FP16 \\
FP8 NVIDIA~\cite{micikevicius2022fp8formatsdeeplearning}
                              & FP32 & FP32 & FP32  & FP16 \\
FP8 IBM / HFP8~\cite{sun2019hybrid}
                              & FP16 & FP16 & FP16  & FP8 \\
INT8 fwd/bwd~\cite{jin2022f8netfixedpoint8bitmultiplication}
                              & FP32 & FP32 & FP32  & FP32  \\
\cite{ghaffari2023integerarithmeticdeeplearning}
                              & FP32 & FP32 & INT16 & FP32  \\
\midrule
\textbf{Ours}
                              & \textbf{INT8} & \textbf{INT8} & \textbf{INT8} & \textbf{INT8} \\
\bottomrule
\end{tabular}
\end{table}

\section{Method}
\label{sec:method}

\begin{figure}[t]
  \centering
  \begin{subfigure}[t]{0.49\textwidth}
    \centering
    \includegraphics[width=\textwidth]{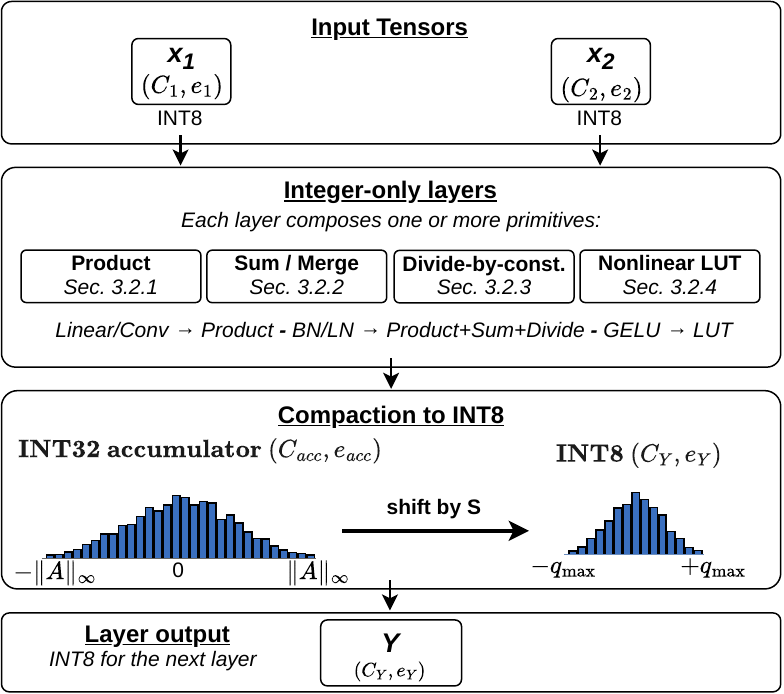}
    \caption{Integer-only layers. Each layer composes one or more integer primitives (product, sum/merge, divide-by-constant, nonlinear LUT) on INT8 inputs into an INT32 accumulator, then compacts back to INT8 via a single shift $S$ with eventual stochastic rounding.}
    \label{fig:overview:layer}
  \end{subfigure}
  \hfill
  \begin{subfigure}[t]{0.49\textwidth}
    \centering
    \includegraphics[width=\textwidth]{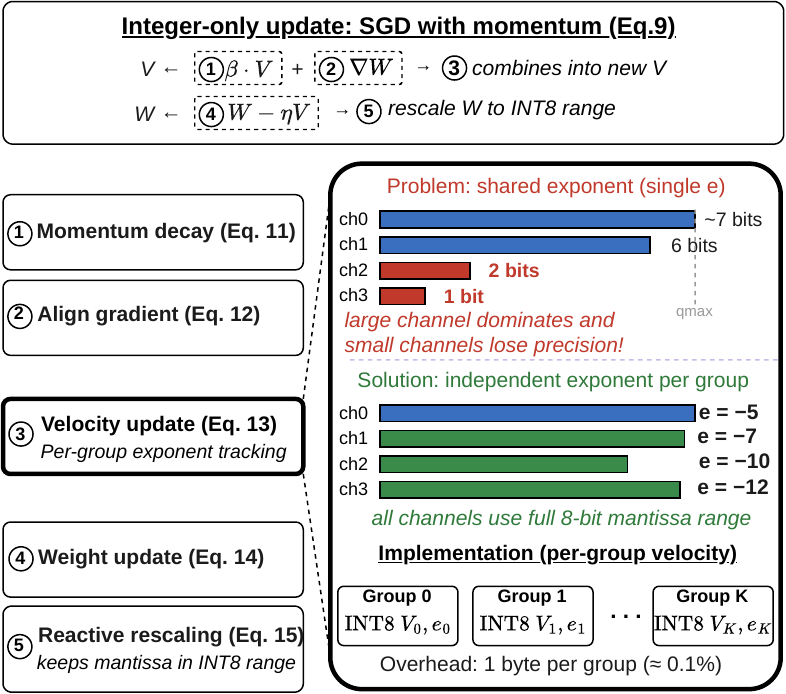}
    \caption{Integer-only weight update (\Cref{eq:sgd_momentum}). Five
    integer steps replace the float SGD-with-momentum step. Per-group velocity exponents let every channel
    use the full INT8 range at $<0.1\%$ overhead (green), while a single exponent would let the largest-gradient
    channel dominate (red).}
    \label{fig:overview:pergroup}
  \end{subfigure}
  \caption{\textbf{Method overview.} All training tensors are integer
  $(C, e)$ pairs (see Eq.~\eqref{eq:sym_tensor_repr}) with no floating-point storage or arithmetic at any stage.
  \textbf{(a)}~Forward and backward of every layer.
  \textbf{(b)}~Integer-only SGD with momentum and per-group velocity.}
  \label{fig:overview}
\end{figure}

In our framework, all persistent tensors (weights, gradients, velocity, saved activations) are stored as INT8 mantissas with power-of-two exponents (\Cref{sec:repr}). Arithmetic operations (inner products, sums) accumulate in transient INT32 accumulators and are immediately compacted back to INT8 by applying a tensor range dependent shift. Whenever this shift moves bits to the right, we use stochastic rounding so that the compacted value remains unbiased while still using the available integer range effectively (\Cref{fig:overview:layer}). This representation lets standard layer operations such as multiplication, sum, division, and nonlinear transformations be implemented directly on mantissas, while the corresponding exponents are updated separately through addition (\Cref{sec:op_layers}).

For weight updates, our framework combines two mechanisms to maintain numerical efficiency and precision. First, \emph{per-group velocity} assigns a separate exponent to each output neuron or channel, which improves precision when gradient magnitudes differ across channels (\Cref{fig:overview:pergroup} and \Cref{sec:op_update}). Second, \emph{reactive rescaling} adjusts the weight exponent after every update step so that the mantissa stays well matched to the available intB range, avoiding both overflow and under-utilization. \Cref{fig:overview} summarizes the full framework.

\subsection{Mantissa-exponent tensor representation}
\label{sec:repr}

For a bit-width $B$ we define $q_{\max} = 2^{B-1}-1$ and store every tensor as a pair $(C, e)$, where $C\in[-q_{\max},q_{\max}]^{\text{shape}}$ is a signed intB mantissa and $e\in\mathbb{Z}$ is a shared exponent ($q_{\max}=127$ for INT8).
The real-valued tensor that this represents is
\begin{equation}
C\cdot2^{e}.
\label{eq:sym_tensor_repr}
\end{equation}
Because the exponent is a power of two, multiplying or dividing by $2^p$ with $p\in\mathbb{N}$ is an exact integer shift on $e$, so no rounding is needed by scale changes that happen to be powers of two.

\paragraph{Compaction.}
Each layer accumulates its output in an INT32 accumulator $A$ and then stores it back in intB using a tensor range dependent shift $S$:
\begin{equation}
C_{out} = \textsc{clip}\bigl( \textsc{shift\_round}(A, S),\ -q_{\max},\ q_{\max} \bigr),
\qquad
e_{out} = e_{in} + S.
\label{eq:storeback_tensor}
\end{equation}
The shift is chosen so that the stored mantissa uses the available INT8 range as fully as possible without exceeding it. Let $m = \lVert A\rVert_\infty$ be the maximum magnitude in $A$. The shift is calculated as
\begin{equation}
S = \bigl\lceil \log_2(m/q_{\max}) \bigr\rceil,
\label{eq:compact_tensor}
\end{equation}
positive when $A$ must be contracted and negative when it can be expanded. In practice we evaluate~$S$ as $\hat S = \mathrm{bit\_length}(m) - \mathrm{bit\_length}(q_{\max}) + 1$, where $\mathrm{bit\_length}(\cdot)$ is a count-leading-zeros instruction and $\mathrm{bit\_length}(q_{\max})$ is a compile-time constant.

\paragraph{Stochastic rounding.}
Right shifts are the only lossy step in the entire framework.
To prevent small but nonzero gradient updates from systematically rounding to zero (which would stall learning), every right shift during training uses stochastic rounding\footnote{Training ResNet-18 on CIFAR-10 with our algorithm and deterministic rounding achieves only 89.3\%, a 5.4\% gap compared to the stochastic rounding version (94.7\%), confirming the advantage of stochastic rounding into the integer-only training pipeline.}:
\begin{equation}
\textsc{shift\_round}(x,S)=\left\lfloor\frac{x+U}{2^S}\right\rfloor,
\qquad
U\sim \mathrm{Unif}\{0,1,\ldots,2^S-1\},
\label{eq:sr_tensor}
\end{equation}
where $\lfloor\cdot\rfloor$ is the floor (rounding toward $-\infty$). The estimator is unbiased for all signed $x$ \cite{ghaffari2023integerarithmeticdeeplearning}. 


\subsection{Layer building blocks}
\label{sec:op_layers}

Given the representation above, the forward and backward passes of all standard layers reduce to four primitives on integer mantissas, with exponents propagated by addition: products, sums, divide-by-constant, and nonlinear lookup tables. The first three (\Cref{sec:op_prod,sec:op_sum,sec:op_div}) are arithmetic; the fourth (\Cref{sec:op_nonlinear}) tabulates each elementwise nonlinearity once at initialization, since under the $(C,e)$ representation each input $(x_{\mathrm{int}}, e_x)$ takes only $2^B N_e$ distinct values, with $N_e\approx 30$ exponents reached in practice. 
\Cref{app:layers} works through how these primitives compose into the layers of our models, namely convolutions, linear layers, BN/LN, the ViT attention block, and the output head.

\subsubsection{Products}
\label{sec:op_prod}
Let $(X,e_X)$ and $(W,e_W)$ be an activation and weight tensor. An inner-dimension-$K$ dot product is performed on the mantissas in an INT32 accumulator, with the exponent obtained by addition:
\begin{equation}
y_{\mathrm{acc}} = \sum_{i=1}^{K} X_i\, W_i,
\qquad
e_{\mathrm{acc}} = e_X + e_W,
\label{eq:prod_acc}
\end{equation}
so that $(y_{\mathrm{acc}}, e_{\mathrm{acc}})$ represents the real value $y_{\mathrm{real}} = y_{\mathrm{acc}}\cdot 2^{e_{\mathrm{acc}}}$. Compacting via \Cref{eq:compact_tensor} yields $(C_Y, e_Y)$ with $e_Y = e_{\mathrm{acc}} + S$. The backward products $\nabla X_{\mathrm{acc}}=GW$ and $\nabla W_{\mathrm{acc}}=G^\top X$ are computed in the same way.
\subsubsection{Sums}
\label{sec:op_sum}
Residual connections and gradient stream merges add two tensors $(C_1,e_1)$ and $(C_2,e_2)$, which requires aligning them to a common exponent. Assume $e_1 \ge e_2$ and let $\Delta e = e_1 - e_2$. When $\Delta e \le \Delta e_{\max}$ we left-shift $C_1$ by $\Delta e$ (exact, no rounding) and set $e_{\text{common}}=e_2$; otherwise we right-shift $C_2$ by $\Delta e$ with stochastic rounding and set $e_{\text{common}}=e_1$. The cap $\Delta e_{\max}$ prevents the left shift from overflowing the wide accumulator. The aligned mantissas $(C'_1,C'_2)$ are summed and accumulated in INT32 and compacted back to intB:
\begin{equation}
C_{\text{out}}=\textsc{clip}\!\left(\textsc{shift\_round}(C'_1+C'_2,\,S),\ -q_{\max},\ q_{\max}\right),
\qquad
e_{\text{out}} = e_{\text{common}} + S.
\label{eq:sum_align}
\end{equation}

\subsubsection{Divide-by-constant}
\label{sec:op_div}
Average pooling and normalization statistics divide a tensor $(C, e)$ by a positive integer constant~$D$.  When $D = 2^k$ with $k \in \mathbb{N}$ is a power of two, the division is absorbed into the exponent at no cost:
\begin{equation}
(C,\, e) \;/\; 2^k \;=\; (C,\; e - k) .
\label{eq:div_pow2}
\end{equation}
For general $D$, we precompute the integer reciprocal $R[D] = \lfloor 2^{S_R} / D \rfloor$, so that $R[D]\cdot D \approx 2^{S_R}$ and the division becomes a multiply followed by a shift by $S_R$:
\begin{equation}
(C,\, e) \;/\; D \;\approx\; \bigl(\textsc{shift\_round}(C \cdot R[D],\; S_R),\; e \bigr).
\label{eq:div_reciprocal}
\end{equation}

\subsubsection{Nonlinear lookup tables}
\label{sec:op_nonlinear}
Each nonlinearity is a finite map tabulated once at initialization and evaluated at runtime by an indexed lookup plus a handful of integer ops. The tables are shared across all instances of a given operator and require no polynomial approximation or floating-point fallback.

\paragraph{Softmax.}
Softmax turns logits into a probability distribution via $\mathrm{softmax}(x_i)=\exp(x_i)/\sum_j \exp(x_j),$ which we realize in three integer steps: (i) 
calculate $x'_i = x_i - \max_i(x_i)$ so that $\exp(x'_i)\in(0,1] \,\,\,\forall\,\,\, i$; (ii) read each $\exp(x'_i)$ from an INT8 LUT; (iii) normalize by an integer reciprocal of the row sum (applied as in Eq.~\eqref{eq:div_reciprocal}). The probabilities are stochastically requantized to unsigned INT8 and cached for the backward pass, which takes the upstream gradient $\partial y$ from the next layer and evaluates the softmax Jacobian $\partial x_i = p_i(\partial y_i - \sum_j \partial y_j\,p_j)$ on the cached probabilities~$p_i$. When followed by cross-entropy, the loss-gradient is $\partial\mathcal{L}/\partial X_L = \tfrac{1}{N}(\mathbf{p} - \mathbf{1}_{y})$, so backpropagation starts from an integer tensor with no need of float fallback for the loss.

\paragraph{GELU.}
We tabulate $\mathrm{GELU}(x)=x\,\Phi(x)$ (where $\Phi$ is the cumulative distribution function for the Gaussian distribution) and its derivative into a pair of INT8 LUTs indexed by $(x_{\mathrm{int}}, e_x)$. Forward is one lookup; backward is one lookup, one integer multiply, and one shift. 

\subsection{Integer-only weight update with scale alignment}
\label{sec:op_update}

Our target update is SGD with momentum applied to each weight tensor $W$, using learning rate $\eta$, momentum coefficient $\beta\in[0,1)$, a persistent \emph{velocity} (momentum buffer) $V$ of the same shape as $W$, and the loss gradient $\nabla W$ produced by backpropagation. Each step applies
\begin{equation}
V \leftarrow \beta V + \nabla W,
\qquad
W \leftarrow W - \eta V.
\label{eq:sgd_momentum}
\end{equation}
Realizing \Cref{eq:sgd_momentum} entirely on integer mantissa-exponent pairs exposes two scale-alignment problems that are much less pronounced in the forward and backward passes. First, gradient magnitudes can vary substantially across output channels within the same weight tensor; with a single shared exponent, large-gradient channels dominate the available dynamic range and updates for smaller channels are heavily quantized or vanish entirely. Second, after many updates the integer weight mantissa drifts: $\|W\|_\infty$ may overflow INT8, or shrink so that most of the available range goes unused, in either case losing precision at the next subtraction. We address the first with per-group velocity (\Cref{sec:pergroup}) and the second with reactive rescaling (\Cref{sec:rescaling}).

\subsubsection{Per-group velocity}
\label{sec:pergroup}

Rather than storing the velocity $V$ as one tensor with a single shared exponent, we partition it along the output-neurons/channels axis of the weights $W$ into $N_g$ groups, each with its own exponent:
\begin{equation}
(V_k,\, e_{V,k}),\qquad V_k\in[-q_{\max},q_{\max}]^{\mathrm{shape}(W_k)},\qquad k=1,\dots,N_g,
\label{eq:vel_repr}
\end{equation}
where $W_k$ denotes the corresponding slice of the weight tensor; the weights themselves keep a single shared exponent $e_W$, since their per-tensor scale changes only slowly across steps (\Cref{sec:rescaling}). Each slice $\nabla W_k$ (along the same axis as the velocity) of the gradient $\nabla W$ is compacted via Eq.~\eqref{eq:storeback_tensor} with its own shift $S_k$ read from $\|\nabla W_k\|_\infty$, yielding $(G_k, e_{G,k})$ with $e_{G,k}=e_{\nabla W}+S_k$. Both $V_k$ and $G_k$ therefore use the full INT8 range regardless of the scales of other slices. The storage overhead is one exponent per velocity slice, below $0.1\%$ for a typical fully connected or convolution layer; all velocity mantissas are INT8 in our experiments.

\subsubsection{Update algorithm}
\label{sec:update_algo}

We encode the scalar hyperparameters as integer (num, shift) pairs: $\beta \approx \beta_{\text{num}}\,2^{-\beta_{\text{shift}}}$ (e.g., $\beta{=}0.9\approx 230/2^8$) and $\eta \approx \eta_{\text{num}}\,2^{-\eta_{\text{shift}}}$, precomputing a sequence $(\eta_{\text{num}}^{(t)}, \eta_{\text{shift}}^{(t)})$ per iteration $t$ to realize cosine, step, or warmup schedules (\Cref{app:recipe}). All arithmetic below is applied independently per group~$k$.

We first apply momentum decay to the previous velocity:
\begin{equation}
D_k = \textsc{shift\_round}(\beta_{\text{num}}\cdot V_k,\;\beta_{\text{shift}}),
\qquad e_{D,k}=e_{V,k}.
\label{eq:momentum_decay}
\end{equation}
We then align the grouped gradient $(G_k, e_{G,k})$ to the velocity exponent $e_{V,k}$. With $\delta_k=e_{G,k}-e_{V,k}$, the aligned gradient is
\begin{equation}
\tilde{G}_k = \textsc{shift\_round}(G_k,\, -\delta_k).
\label{eq:grad_align}
\end{equation}

The new velocity is obtained by accumulation followed by compaction back to INT8:
\begin{equation}
V_k \leftarrow \textsc{clip}\bigl(\textsc{shift\_round}(D_k + \tilde G_k,\; S_k'),\; -q_{\max},\; q_{\max}\bigr),
\qquad
e_{V,k} \leftarrow e_{V,k} + S_k',
\label{eq:vel_compact}
\end{equation}
with $S_k'$ chosen by \Cref{eq:compact_tensor} from $\|D_k+\tilde G_k\|_\infty$.
Finally, we form the update direction $\eta_{\text{num}}\!\cdot\! V_k$ at exponent $-\eta_{\text{shift}}+e_{V,k}$, align it to the weight exponent $e_W$ (as in \Cref{eq:grad_align}), and subtract from the current weights in a single shift-round:
\begin{equation}
W_k \leftarrow W_k - \textsc{shift\_round}\bigl(\eta_{\text{num}} \cdot V_k,\; e_W + \eta_{\text{shift}} - e_{V,k}\bigr).
\label{eq:weight_step}
\end{equation}
Weight decay is handled the same way as the momentum coefficient: $\lambda$ is stored as $(\lambda_{\text{num}}, \lambda_{\text{shift}})$ and~$\lambda W_k$ is added to $G_k$ before the velocity update (see \Cref{app:recipe} for details).

\subsubsection{Reactive rescaling}
\label{sec:rescaling}

After the groupwise updates have been applied, the integer weight tensor may either exceed the INT8 representable range or use only a small fraction of it. Letting $m=\|W\|_\infty$, a single signed shift restores the mantissa to the band $[q_{\max}/4,\,q_{\max}]$:
\begin{equation}
\begin{aligned}
W &\leftarrow \textsc{shift\_round}(W,\,S), \\
e_W &\leftarrow e_W + S,
\end{aligned}
\qquad
S = \begin{cases}
\bigl\lceil\log_2(m/q_{\max})\bigr\rceil, & m > q_{\max}, \\[1pt]
-\bigl\lfloor\log_2\!\bigl(q_{\max}/(4m)\bigr)\bigr\rfloor, & 0 < m < q_{\max}/4, \\[1pt]
0, & \text{otherwise}.
\end{cases}
\end{equation}
The first case ($S>0$) rounds the overflow mantissa down by $S$ bits, keeping the represented real value within one LSB; the second case ($S\le 0$) is an exact left shift. 

\section{Experiments}
\label{sec:results}

\paragraph{Setup.}
We evaluate three architecture classes across three datasets: ResNet~\cite{he2016deep} on CIFAR-10, CIFAR-100~\cite{krizhevsky2009learning}, and ImageNet~\cite{deng2009imagenet}; MobileNetV2~\cite{sandler2018mobilenetv2} on CIFAR-10; ViT-Ti from scratch on CIFAR-10; and ViT-B~\cite{dosovitskiy2021an} fine-tuned from a pretrained checkpoint on CIFAR-10. All runs use SGD with momentum $\beta{=}0.9$ (encoded as $(\beta_{\text{num}}, \beta_{\text{shift}}){=}(230, 8)$ per \Cref{sec:update_algo}) and a cosine LR schedule with linear warmup, precomputed as $(\eta_{\text{num}}, \eta_{\text{shift}})$ pairs. The default configuration uses INT8 for activations, weights, gradients, and velocity, with INT32 accumulators. Full per-model hyperparameters, augmentation, input quantization, and the integer LR schedule are listed in \Cref{app:recipe}.

\begin{figure}[htbp]
\centering
\includegraphics[width=\textwidth]{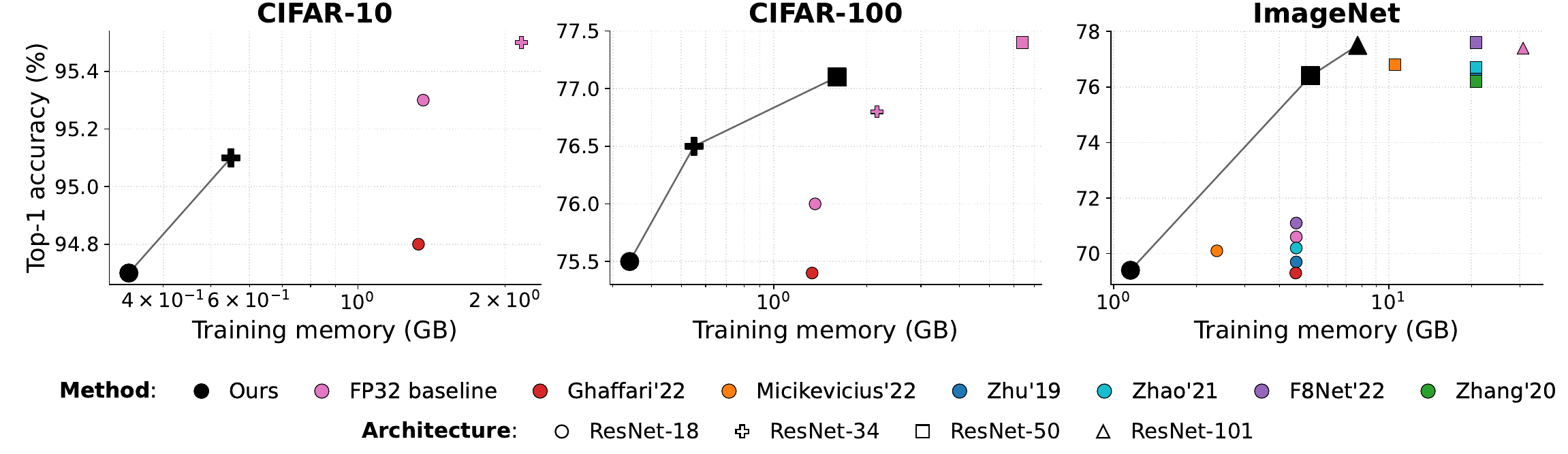}
\caption{Accuracy vs training memory on CIFAR-10, CIFAR-100 and ImageNet. Each point is one (method, model) entry from \Cref{tab:main_results}.
}
\label{fig:pareto_imagenet}
\end{figure}

\paragraph{Main results.}
\Cref{tab:main_results} summarizes the comparison. Ours is the only method that holds every persistent tensor in the training pipeline at INT8 (weights, gradients, velocity, and saved activations), yielding a $4\times$ reduction in training memory across every configuration while staying within a few percent of FP32 accuracy and within $0.1$--$0.7\%$ of \citet{ghaffari2023integerarithmeticdeeplearning}, despite eliminating their FP32 master weights, FP32 optimizer state, and float softmax fallback. 
The memory gap to other low-precision methods is structural: training memory is dominated by saved-for-backward activations~\cite{chen2016sublinear} ($80\text{--}99\%$ of the FP32 footprint across our configurations, per-config breakdown in \Cref{app:memory}), and prior INT8 forward/backward methods leave activations in FP32 between layers. The Pareto front corresponding to each method in~\Cref{tab:main_results} is reported in \Cref{fig:pareto_imagenet}. Our algorithm is particularly advantageous at low memory budgets: e.g., on ImageNet it reaches $76.4\%$ at $5.20$\,GB memory on ResNet-50, matching the FP32 ResNet-18 ($4.61$\,GB, $70.6\%$) memory footprint within~$13\%$ with an increase of $+5.8\%$ in accuracy.

\paragraph{CNN results.}
On ResNet-18 we trail FP32 by $0.6\%$ on CIFAR-10 ($94.7$ vs.\ $95.3$), $0.5\%$ on CIFAR-100 ($75.5$ vs.\ $76.0$), and $1.2\%$ on ImageNet ($69.4$ vs.\ $70.6$), at roughly $4{\times}$ smaller training-memory footprint than the baselines. 
On ResNet-50/ImageNet our algorithm reaches $76.4\%$, a $0.3\%$ gap from FP32 ($76.7\%$) and within $1.2\%$ of the best INT8 forward/backward result (\citet{jin2022f8netfixedpoint8bitmultiplication}, $77.6\%$), while running at $5.20$ vs.\ $20.82$\,GB (same $4{\times}$ memory reduction).
MobileNetV2/CIFAR-10 widens to $-3.2\%$ ($91.5$ vs.\ $94.7$), reflecting the architecture known sensitivity to quantization. Two factor compound: depthwise separable convolutions decompose each spatial filter into per-channel kernels with very few weights, making each channel gradient signal small and quantization-sensitive, even with our per-output-channel velocity (which contributes +2.2\%, see \Cref{tab:pergroup_ablation}); and the inverted-residual blocks introduce a high density of sequential requantization points across depth. 
To our knowledge this is the first fully integer training of ResNet and MobileNet architectures.

\begin{table}[h!]
\centering
\caption{Test accuracy (\%) and training memory (GB) on CIFAR-10, CIFAR-100 and ImageNet. Training details and hyperparameters are reported in \Cref{app:recipe}, while the memory computation is detailed in \Cref{app:memory}.}
\label{tab:main_results}
\small
\setlength{\tabcolsep}{4pt}
\begin{tabular}{lllccl}
\toprule
\multirow{2}{*}{\textbf{Dataset}}
  & \multirow{2}{*}{\textbf{Model}}
  & \multirow{2}{*}{\textbf{Method}}
  & \multirow{2}{*}{\textbf{Acc.\ (\%)}}
  & \textbf{Training}
  & \multirow{2}{*}{\textbf{Precision}} \\
  & & & & \textbf{mem. (GB)} & \\
\midrule
\multirow{14}{*}{CIFAR-10}
  & \multirow{3}{*}{ResNet-18}
  & FP32~\cite{ghaffari2023integerarithmeticdeeplearning}
                            & 95.3 & 1.36 & FP32 \\
  & & \cite{ghaffari2023integerarithmeticdeeplearning}
                            & 94.8 & 1.33 & INT8/16 + FP32 \\
  & & Ours & 94.7 & $0.34$ & INT8 \\
\cmidrule(l){2-6}
  & \multirow{2}{*}{ResNet-34}
  & FP32~\cite{mbsariyildiz_resnet_pytorch}
                            & 95.5 & 2.16 & FP32 \\
  & & Ours & 95.1 & $0.55$ & INT8 \\
\cmidrule(l){2-6}
  & \multirow{4}{*}{MobileNetV2}
  & FP32~\cite{zhao2021distributionadaptiveint8quantization}
                            & 94.7 & 1.82 & FP32 \\
  && \cite{zhu2019unifiedint8trainingconvolutional}
                            & 93.3 & 1.82 & INT8 + FP32 \\
  && \cite{zhao2021distributionadaptiveint8quantization}
                            & 94.3 & 1.82 & INT8 + FP32 \\
  & & Ours & 91.5 & $0.45$ & INT8 \\
\cmidrule(l){2-6}
  & \multirow{2}{*}{ViT-Ti}
  & FP32~\cite{aleptonic_vit_scratch}
                            & 90.9 & 7.60 & FP32 \\
  & & Ours & 88.0 & $1.90$ & INT8 \\
\cmidrule(l){2-6}
  & \multirow{3}{*}{ViT-B}
  & FP32~\cite{ghaffari2023integerarithmeticdeeplearning}
                            & 99.1 & 9.33 & FP32 \\
  & & \cite{ghaffari2023integerarithmeticdeeplearning}
                            & 98.8 & 9.17 & INT8/16 + FP32 \\
  &(fine-tune)& Ours & 96.1 & $2.33$ & INT8 \\
\midrule
\multirow{7}{*}{CIFAR-100}
  & \multirow{3}{*}{ResNet-18}
  & FP32~\cite{ghaffari2023integerarithmeticdeeplearning}
                            & 76.0 & 1.36 & FP32 \\
  & & \cite{ghaffari2023integerarithmeticdeeplearning}
                            & 75.4 & 1.33 & INT8/16 + FP32 \\
  & & Ours & 75.5 & $0.34$ & INT8 \\
\cmidrule(l){2-6}
  & \multirow{2}{*}{ResNet-34}
  & FP32~\cite{weiaicunzai_cifar100}
                            & 76.8 & 2.16 & FP32 \\
  & & Ours & 76.5 & $0.55$ & INT8 \\

\cmidrule(l){2-6}
  & \multirow{2}{*}{ResNet-50}
  & FP32~\cite{weiaicunzai_cifar100}
                            & 77.4 & 6.40 & FP32 \\
  & & Ours & 77.1 & $1.60$ & INT8 \\
\midrule
\multirow{16}{*}{ImageNet}
  & \multirow{7}{*}{ResNet-18}
  & FP32~\cite{micikevicius2022fp8formatsdeeplearning}
                            & 70.6 & 4.61 & FP32 \\
  && \cite{micikevicius2022fp8formatsdeeplearning}
                            & 70.1 & 2.37 & FP8 + FP32 \\
  && \cite{zhu2019unifiedint8trainingconvolutional}
                            & 69.7 & 4.61 & INT8 + FP32 \\
  && \cite{zhao2021distributionadaptiveint8quantization}
                            & 70.2 & 4.61 & INT8 + FP32 \\
  && \cite{jin2022f8netfixedpoint8bitmultiplication}
                            & 71.1 & 4.61 & INT8 + FP32 \\
  & & \cite{ghaffari2023integerarithmeticdeeplearning}
                            & 69.3 & 4.59 & INT8/16 + FP32 \\
  & & Ours & 69.4 & $1.15$ & INT8 \\
\cmidrule(l){2-6}
  & \multirow{7}{*}{ResNet-50}
  & FP32~\cite{micikevicius2022fp8formatsdeeplearning}
                            & 76.7 & 20.82 & FP32 \\
  && \cite{micikevicius2022fp8formatsdeeplearning}
                            & 76.8 & 10.55 & FP8 + FP32 \\
  && \cite{zhu2019unifiedint8trainingconvolutional}
                            & 76.3 & 20.82 & INT8 + FP32 \\
  && \cite{9157439}
                            & 76.2 & 20.82 & INT8 + FP32 \\
  && \cite{zhao2021distributionadaptiveint8quantization}
                            & 76.7 & 20.82 & INT8 + FP32 \\
  && \cite{jin2022f8netfixedpoint8bitmultiplication}
                            & 77.6 & 20.82 & INT8 + FP32 \\
  & & Ours & 76.4 & $5.20$ & INT8 \\
\cmidrule(l){2-6}
  & \multirow{2}{*}{ResNet-101}
  & FP32~\cite{torchvision_models}
                            & 77.4 & 30.84 & FP32 \\
  & & Ours & 77.5 & $7.71$ & INT8 \\
\bottomrule
\end{tabular}
\end{table}
\paragraph{Vision Transformers.}
ViT-Ti from scratch on CIFAR-10 reaches $88.0\%$, a $-2.9\%$ gap from FP32 ($90.9\%$). ViT-B fine-tuning on CIFAR-10 reaches $96.1\%$ in 50 epochs, a $-3.0\%$ gap from FP32 ($99.1\%$). To our knowledge, this is the first fully integer-only training or fine-tuning of a Vision Transformer: every layer (attention softmax, $1/\sqrt{d_h}$ scaling, LayerNorm, and GELU) runs on integer mantissa-exponent pairs with no floating-point fallback at any stage, (see~\Cref{tab:vit_ablation} for the accuracy impact of quantizing each layer). The closest prior point, \citet{ghaffari2023integerarithmeticdeeplearning}, retains FP32 master weights, an FP32 optimizer, and a float softmax in attention. ViTs are typically trained with Adam/AdamW, while our framework supports only SGD+momentum at the moment; extending the integer pipeline to adaptive optimizers could improve performance and is a natural next step, since their moment updates require integer square roots and divisions that complicate the scale bookkeeping.

\paragraph{Lower-bit regimes.}
\Cref{tab:low_bit} sweeps the bit-width B below INT8 on ResNet-18/CIFAR-10 on our algorithm and on the algorithm from~\cite{ghaffari2023integerarithmeticdeeplearning}. Both methods retain their default optimizer state while weights and activations drops to B bits: our velocity stays at INT8, while \citet{ghaffari2023integerarithmeticdeeplearning} master weights stay at FP32 and their velocity at INT16. Notably, our algorithm retains $70.5\%$ at $B{=}4$ where~\cite{ghaffari2023integerarithmeticdeeplearning} diverges. Memory scales with B for ours ($0.34{\to}0.17$\,GB), while~\cite{ghaffari2023integerarithmeticdeeplearning} retains the persistent FP32 state.
\begin{table}[h!]
\centering
\caption{Accuracy (\%) and training memory (GB) on ResNet-18 / CIFAR-10 (batch~256) across bit-widths $\text{B}\in\{8,7,6,5,4\}$.}
\label{tab:low_bit}
\small
\setlength{\tabcolsep}{4pt}
\begin{tabular}{llcccccl}
\toprule
\textbf{Method} & \textbf{Metric}
  & \textbf{INT8} & \textbf{INT7} & \textbf{INT6} & \textbf{INT5} & \textbf{INT4} &  \textbf{Precision} ($\text{B}\in\{4,\ldots,8\}$) \\
\midrule
    & Acc (\%) & 94.8 & 94.7 & 94.47 & 88.5 & Diverge &  \\
\multirow{-2}{*}{\cite{ghaffari2023integerarithmeticdeeplearning}}
    & Mem (GB) & 1.33 & 1.33 & 1.33 & 1.33 & 1.33 & \multirow{-2}{*}{intB + FP32, velocity INT16} \\
\cmidrule(l){1-8}
    & Acc (\%) & 94.7 & 94.1 & 91.8 & 83.3 & 70.5 &\\
\multirow{-2}{*}{Ours}
    & Mem (GB) & 0.34 & 0.30 & 0.26 & 0.22 & 0.17 & \multirow{-2}{*}{intB, velocity\,INT8} \\
\bottomrule
\end{tabular}
\end{table}

\paragraph{Per-group velocity ablation.}
\Cref{tab:pergroup_ablation} isolates the contribution of per-group velocity (\Cref{sec:pergroup}) on CIFAR-10. Disabling it costs $0.9\%$ on ResNet-18, $2.2\%$ on MobileNetV2, and $2.7\%$ on ViT-Ti. 
\begin{table}[h!]
\centering
\caption{Effect of per-group velocity on test accuracy (\%) and exponent memory.}
\label{tab:pergroup_ablation}
\small
\setlength{\tabcolsep}{3pt}
\begin{tabular}{llccccc}
\toprule
  \multirow{2}{*}{\textbf{Dataset}}
  & \multirow{2}{*}{\textbf{Model}}
  & \textbf{FP32}
  & \multicolumn{2}{c}{\textbf{w/o per-group velocity}}
  & \multicolumn{2}{c}{\textbf{w/ per-group velocity}} \\
\cmidrule(lr){4-5}\cmidrule(lr){6-7}
  & & \textbf{(Reference)}
  & \textbf{Acc.} & \textbf{Exp.\ mem.}
  & \textbf{Acc.} & \textbf{Exp.\ mem.} \\
\midrule
\multirow{3}{*}{CIFAR-10}
  & ResNet-18    & 95.3 & 93.8 ($-1.5$) & $0.02$ KB
                        & 94.7 ($-0.6$) & $4.7$ KB  \\
  & MobileNetV2  & 94.7 & 89.3 ($-5.4$) & $0.05$ KB
                        & 91.5 ($-3.2$) & $16.7$ KB \\
  & ViT-Ti       & 90.9 & 85.3 ($-5.6$) & $0.05$ KB
                        & 88.0 ($-2.9$) & $20.5$ KB \\
\bottomrule
\end{tabular}
\end{table}



\section{Conclusion}
\label{sec:conclusion}

We demonstrated that a full SGD-with-momentum training pipeline can be realized end to end on INT8 mantissa-exponent tensors with no floating-point fallback at any stage. The resulting algorithm is, to our knowledge, the first integer-only training framework that scales to standard computer-vision benchmarks at modern depth: it trains ResNet-18/34/50/101 and MobileNetV2 to within a few percent of FP32 across CIFAR-10, CIFAR-100, and ImageNet, fine-tunes ViT-B integer-only, and trains ViT-Ti from scratch. Methodologically, per-output-channel velocity exponents are essential for integer optimizer state.
Because every persistent tensor is INT8, training memory is roughly $4\times$ smaller than FP32 across every configuration we tested. This shifts the accuracy-memory frontier: at a fixed memory budget, the integer pipeline trains substantially deeper and more accurate architectures than FP32 can admit; at a fixed architecture, it trains within a few percent of the FP32 baseline at a quarter of the memory.
The same pipeline can in principle run on hardware without floating-point support (e.g. microcontrollers, in-memory compute substrates) opening on-device training of ResNet- and ViT-scale models.  

\paragraph{Limitations and future work.}
Extending the framework to adaptive optimizers (an integer Adam, requiring integer square roots, integer division, and exponent bookkeeping for the bias-correction terms) is the most substantive open direction for moving to architectures that are typically trained with Adam or AdamW. Replacing power-of-two exponents with richer scaling schemes (learned scales, non-power-of-two block exponents) would use dynamic range more efficiently and likely narrow the remaining accuracy gap on MobileNetV2 and ViT. Operator fusion (e.g. reducing the requantization points per ViT block) is a possible engineering direction for requantization-heavy architectures. Finally, bringing the framework to LLM training and parameter-efficient fine-tuning (LoRA) is a natural next step, with the same structural advantage of keeping the optimizer state at INT8 rather than FP32.

\begin{ack}
This work was carried out in the EssilorLuxottica Smart Eyewear Lab, a Joint Research Center between EssilorLuxottica and Politecnico di Milano.
\end{ack}

\bibliography{reference}
\bibliographystyle{plainnat}


\appendix


\section{Integer primitives, shifts, and safety caps}
\label{app:primitives}

This appendix defines the low-level integer primitives used throughout the paper: the clip and signed shift-and-round operators, the stochastic-rounding operator, and the accumulator safety caps that bound the compaction shift of \Cref{eq:compact_tensor}.

\paragraph{Clip and shift-round.}
The two primitives from which every rescale in the paper is built are saturation and signed shift-with-round:
\begin{equation}
\textsc{clip}(x,a,b)=\min(\max(x,a),b),
\qquad
\textsc{shift\_round}(x,S)=
\begin{cases}
\bigl\lfloor (x+b(S))/2^S \bigr\rfloor, & S>0,\\[2pt]
x\cdot 2^{-S}, & S\le 0,
\end{cases}
\label{eq:shift_round_def}
\end{equation}
where $\lfloor\cdot\rfloor$ denotes the floor toward $-\infty$ and $b(S)\ge 0$ is an integer bias that selects the rounding mode. We use $b(S)=U\sim\mathrm{Unif}\{0,\dots,2^S{-}1\}$ during training (stochastic rounding, \Cref{eq:sr_tensor}) and $b(S)=2^{S-1}$ at inference (deterministic midpoint). Left shifts ($S\le 0$) are exact and involve no rounding choice. The compaction shift of \Cref{eq:compact_tensor} uses $\lceil\log_2 m\rceil$, computed by integer bit-length (count-leading-zeros); we handle the edge case $m=\|A\|_\infty=0$ by setting $S=0$.

\paragraph{Stochastic rounding is integer-only.}
Stochastic rounding samples a uniform integer $U$ per tensor element and computes
\begin{equation}
\textsc{shift\_round}(x,S) \;=\; \bigl\lfloor (x+U)/2^S \bigr\rfloor,\qquad U\sim\mathrm{Unif}\{0,\dots,2^S{-}1\},
\label{eq:sr_app}
\end{equation}
so every ingredient is integer: a xorshift or PCG generator produces $U$ in 3--5 integer ops (shifts and xors), and the addition and right shift are native integer instructions, so the same kernel runs on hardware without an FPU. The per-element PRNG cost is negligible against the $K$-wide inner product that produces $x$: $K$ ranges from $9$ for a $3{\times}3$ depthwise convolution to $4608$ for a ViT-B MLP projection, so the PRNG adds well under $1\%$ to the per-layer arithmetic. We hold one PRNG state per tensor to avoid cross-thread contention; at inference $U$ is replaced by the deterministic midpoint $2^{S-1}$, eliminating the PRNG entirely.

\paragraph{Accumulator safety caps.}
The shift chosen by \Cref{eq:compact_tensor} must not cause the next reduction to overflow. Let the wide accumulator be $n$-bit signed with maximum magnitude $A_{\max}=2^{n-1}-1$. For a $K$-wide inner product with $|x_i|,|w_i|\le q_{\max}$ we have $|\sum_{i=1}^{K} x_i w_i|\le K\,q_{\max}^2$, and for an $M$-way sum of aligned mantissas $|\sum_i x_i|\le M\,q_{\max}$. The largest admissible expansions are therefore
\begin{equation}
S_{\mathrm{cap}}^{\mathrm{prod}} = \bigl\lfloor \log_2\!\bigl(A_{\max}/(K\,q_{\max}^2)\bigr) \bigr\rfloor ,
\qquad
S_{\mathrm{cap}}^{\mathrm{sum}} = \bigl\lfloor \log_2\!\bigl(A_{\max}/(M\,q_{\max})\bigr) \bigr\rfloor .
\label{eq:caps}
\end{equation}
If a cap is negative the reduction must be tiled, the operands pre-shifted, or a wider accumulator used. With INT32 accumulators ($A_{\max}\approx 2.1\cdot 10^9$) and $q_{\max}=127$ (INT8), both caps stay comfortably positive for every layer size in our models.

\section{Concrete layer instantiations}
\label{app:layers}

Each layer is built by composing the primitives of \Cref{sec:op_layers,sec:op_update}. \Cref{tab:layer_summary} maps each layer or operator used in our models to the primitives it reuses and any layer-specific lookup table; the subsections below fill in operator-specific details (LUT definitions, backward equations, and architectural constants). 
Residual adds and backward gradient merges reuse the sum operator directly (\Cref{eq:sum_align}), and average pooling reuses the divide-by-constant primitive (\Cref{eq:div_pow2,eq:div_reciprocal}).

\begin{table}[h]
\centering
\caption{Layer and operator instantiations used by our framework.}
\label{tab:layer_summary}
\small
\setlength{\tabcolsep}{4pt}
\begin{tabular}{llc}
\toprule
\textbf{Layer / operator} & \textbf{Primitives used} & \textbf{Ref.} \\
\midrule
Convolution (incl.\ 1$\times$1, depthwise) & Product                              & \Cref{eq:prod_acc,eq:compact_tensor} \\
Linear / fully-connected                   & Product                              & \Cref{eq:prod_acc,eq:compact_tensor} \\
Patch embedding (strided conv)             & Product                              & \Cref{eq:prod_acc,eq:compact_tensor} \\
Residual add, gradient merge               & Sum                                  & \Cref{eq:sum_align} \\
Average pool ($K{\times}K$)                & Divide-by-constant                   & \Cref{eq:div_pow2,eq:div_reciprocal} \\
ReLU / ReLU6                               & Clip                                 & \Cref{app:activations} \\
GELU                                       & Product, LUT                         & \Cref{app:activations} \\
Batch / Layer normalization                & Product, sum, divide-by-constant     & \Cref{app:batchnorm} \\
Softmax                                    & Product, divide-by-constant, LUT     & \Cref{app:poly_softmax} \\
Cross-entropy loss                         & Softmax, divide-by-constant, LUT     & \Cref{app:ce} \\
ViT attention (Q/K/V, score, context)      & Product, softmax, constant rescale   & \Cref{app:vit} \\
\bottomrule
\end{tabular}
\end{table}

\subsection{Activations}
\label{app:activations}

Our models use three pointwise activations: ReLU (ResNet-18), ReLU6 (MobileNetV2), and GELU (ViT-Ti/B). \Cref{tab:activations} compares their integer forward and backward; none changes the exponent $e$, and only GELU requires a lookup table.

\begin{table}[h]
\centering
\caption{Pointwise activations used in our models. All operations are integer on the int$B$ mantissa $C$; the exponent $e$ is carried through unchanged. $dY$ denotes the upstream gradient.}
\label{tab:activations}
\small
\setlength{\tabcolsep}{4pt}
\begin{tabular}{lllc}
\toprule
\textbf{Activation} & \textbf{Forward (on $C$)} & \textbf{Backward (grad)}& \textbf{LUT} \\
\midrule
ReLU                & $\max(C,0)$                                                                            & $dY\cdot\mathbf{1}[C>0]$                                        & ---          \\
ReLU6                  & $\textsc{clip}(C,0,c),\;c=\min(6\gg e,q_{\max})$                                      & $dY\cdot\mathbf{1}[0<C<c]$                                      & ---          \\
GELU                       & $\textsc{LUT}_{\mathrm{fwd}}[e, C]$ (\Cref{eq:gelu_luts})                              & $(dY\cdot\textsc{LUT}_{\mathrm{bwd}}[e, C])\gg B_G$              & INT8 $[30,256]$ \\
\bottomrule
\end{tabular}
\end{table}

ReLU and ReLU6 are single-clip operations with no LUT. For ReLU6, expressing the clipping threshold~$6$ in the mantissa domain requires the per-exponent cap $c=\min(6\gg e, q_{\max})$ (a plain signed shift: right by $e$ for $e>0$, left by $|e|$ for $e\le 0$; no rounding is needed because $c$ is a deterministic bound, and truncation only shrinks the clip region). When $6\gg e\ge q_{\max}$ the clip reduces to ReLU because the mantissa already saturates below~$6$.

\paragraph{GELU LUTs.}
Because GELU is pointwise, its entire INT8 behavior fits in a pair of INT8 tables indexed by $(e_x, x_{\mathrm{int}})$:
\begin{equation}
\begin{aligned}
\textsc{LUT}_{\mathrm{fwd}}[e_x,x]
&= \textsc{clip}\!\bigl(\textsc{round}(\mathrm{GELU}(x\cdot 2^{e_x})/2^{e_x}),\, -q_{\max},\, q_{\max}\bigr), \\
\textsc{LUT}_{\mathrm{bwd}}[e_x,x]
&= \textsc{round}\!\bigl(\mathrm{GELU}'(x\cdot 2^{e_x})\cdot 2^{B_G}\bigr),
\end{aligned}
\label{eq:gelu_luts}
\end{equation}
with $B_G{=}6$ so that $\mathrm{GELU}'\in[-0.13,1.13]$ fits in INT8 as $[-8,72]$. Reducing $B_G$ from the common $8$ to $6$ produces no measurable change in backward accuracy versus autograd, because the downstream compaction to INT8 already limits gradient precision to the input bit-width. The backward multiply in \Cref{tab:activations} stays in INT32 since $|\textsc{LUT}_{\mathrm{bwd}}|\le 72$ and $|dY|\le q_{\max}$.

\paragraph{Cross-bit-width LUT reuse.}
For $B<8$, an int$B$ mantissa $x_B$ at exponent $e_B$ equals the INT8 mantissa $x_B\ll(8{-}B)$ at exponent $e_B-(8{-}B)$; a single INT8 GELU table therefore serves all bit widths by shifting the input into the INT8 domain and stochastically shifting the result back.

\subsection{Batch and layer normalization}
\label{app:batchnorm}\label{app:layernorm}

BN and LN share the same integer pipeline; only the reduction axis differs (the batch for BN, the feature dimension for LN). Given an input mantissa $C_x$ at exponent $e_x$ reduced over $D{=}m_D\,2^{s_D}$ elements, the layer computes $y=\gamma\,(x-\mu)/\sqrt{\sigma^2+\epsilon}+b$ in five integer stages. A single reciprocal table $R[m_D]=\lfloor 2^{S_R}/m_D\rfloor$ (we use $S_R{=}30$) is precomputed once and shared by every BN/LN layer at that odd-part $m_D$. Every step below is integer-only: no float intermediate is constructed.

\paragraph{(1) Mean.}
The elementwise sum $T_\mu=\sum_i C_{x,i}$ is the integer sum primitive (\Cref{eq:sum_align}); because the operands already share $e_x$, no alignment is needed. Dividing by $D=m_D\,2^{s_D}$ is a multiply-and-shift through the reciprocal table:
\begin{equation}
C_\mu = \textsc{shift\_round}\!\bigl(T_\mu\cdot R[m_D],\;S_R+s_D\bigr),\qquad e_\mu=e_x.
\label{eq:ln_mean}
\end{equation}

\paragraph{(2) Variance.}
The centered square sum $T_{\sigma^2}=\sum_i (C_{x,i}-C_\mu)^2$ reuses the product-accumulator (\Cref{eq:prod_acc}); the safety cap (\Cref{eq:caps}) covers it because $|C_{x,i}-C_\mu|\le 2q_{\max}$. One more multiply-and-shift by $R[m_D]$ yields the variance mantissa:
\begin{equation}
C_{\sigma^2} = \textsc{shift\_round}\!\bigl(T_{\sigma^2}\cdot R[m_D],\;S_R+s_D\bigr),\qquad e_{\sigma^2}=2e_x.
\label{eq:ln_var}
\end{equation}

\paragraph{(3) Integer square root and reciprocal.}
We take $\sigma=\lfloor\sqrt{C_{\sigma^2}+\epsilon_{\mathrm{int}}}\rfloor$ by the standard Newton integer square-root (a fixed number of integer shifts, adds, and compares). Count-leading-zeros decomposes $\sigma=m_\sigma\,2^{s_\sigma}$, and one INT32 divide per reduction row builds $\mathrm{inv}_\sigma=\lfloor 2^{S_R}/m_\sigma\rfloor$ (the cost amortizes over the $D$ normalizations below); $e_\sigma=e_x$.

\paragraph{(4) Normalize.}
Each centered value is multiplied by $\mathrm{inv}_\sigma$ in an INT32 accumulator and compacted back to int$B$ with stochastic rounding (\Cref{eq:compact_tensor}):
\begin{equation}
(C_{\hat x,i},\,e_{\hat x}) \;=\; \textsc{compact}\!\bigl((C_{x,i}-C_\mu)\cdot \mathrm{inv}_\sigma,\; -S_R-s_\sigma\bigr).
\label{eq:ln_normalize}
\end{equation}
The natural exponent passed to compact cancels both the $2^{S_R}$ in $\mathrm{inv}_\sigma$ and the $2^{s_\sigma}$ in $\sigma$, leaving the real-valued $(x-\mu)/\sigma$ as the interpretation of $(C_{\hat x},e_{\hat x})$.

\paragraph{(5) Affine.}
The learnable $(\gamma,e_\gamma)$ and $(b,e_b)$ are integer parameters updated by the weight-update pipeline of \Cref{sec:op_update}; the final output is formed by one product and one sum: $y=\textsc{sum}\bigl(\textsc{prod}(\hat x,\gamma),\,b\bigr)$ using \Cref{eq:prod_acc,eq:compact_tensor,eq:sum_align}.

The backward pass caches $(C_\mu, C_{\sigma^2}, \mathrm{inv}_\sigma, s_\sigma)$ and back-propagates through the five stages with only the product, sum, and shift-round primitives; no step needs floating point. In summary, the entire BN/LN layer uses integer sums, integer products, a single INT32 divide per reduction row, one integer square root, and shift-rounds.

\paragraph{Running statistics.}
The per-channel running mean and variance used at evaluation are stored as integer mantissa-exponent pairs 
$(\mu^{\mathrm{run}}, e_{\mu^{\mathrm{run}}})$, $(\sigma^{2,\mathrm{run}}, e_{\sigma^{2,\mathrm{run}}})$ 
updated by an exponential moving average 
\begin{equation}
    \mu^{\mathrm{run}} \leftarrow (1-\rho)\,\mu^{\mathrm{run}} + \rho\,\mu.
\end{equation}
The EMA factor $\rho$ is represented as $\rho \approx \rho_{\mathrm{num}}/2^{\rho_{\mathrm{shift}}}$ (e.g.\ $\rho = 0.1 \approx 13/2^7$) and the update uses the same bounded-shift alignment policy as the velocity update, so the running statistics remain integer throughout training and inference. They are not updated through gradient descent, so they do not affect the training-state accounting in \Cref{tab:storage_formats}.

\subsection{Output head: softmax and cross-entropy}
\label{app:output_head}

From the final logits $(X_L, e_{X_L})$ of a batch of $N$ samples over $K$ classes, with $X_L\in\mathbb{Z}^{N\times K}$ and label vector $y\in\{0,\dots,K{-}1\}^N$, the classification head produces three quantities:
\begin{enumerate}
    \item fixed-point softmax probabilities $p\in[0,2^{B}{-}2]^{N\times K}$ stored as unsigned int$B$ ($B{=}8$, so $p\in[0,254]$), used wherever a probability is consumed (e.g.\ the attention scores of \Cref{app:vit});
    \item the scalar cross-entropy loss $\mathcal{L}=-\tfrac{1}{N}\sum_n\log(p_{n,y_n})$, used only for logging;
    \item the loss gradient $\partial\mathcal{L}/\partial X_L=\tfrac{1}{N}(p-\mathbf{1}_{y})$, which seeds the reverse-order layer loop of the training step.
\end{enumerate}
All three share the two precomputed INT8 tables of \Cref{tab:output_head_luts}; every other operation is an integer sum, product, divide-by-constant, or shift-round.

\begin{table}[h]
\centering
\caption{Precomputed INT8 tables for the output head. Built once at initialization and shared across every output instance in the model.}
\label{tab:output_head_luts}
\small
\setlength{\tabcolsep}{5pt}
\begin{tabular}{llllll}
\toprule
\textbf{LUT} & \textbf{Entry} & \textbf{Domain} & \textbf{Scale} & \textbf{Shape} & \textbf{Size} \\
\midrule
$\textsc{LUT}_{\exp}$ & $\textsc{round}\!\bigl(\exp(k\,2^{e_x})\cdot 2^{L}\bigr)$       & $k\in[-255,0]$,\,${\sim}30$ exponents & $L{=}7$      & $[30,256]$ & ${\sim}8$\,KB \\
$\textsc{LUT}_{\log}$ & $\textsc{round}\!\bigl(-\log(p/2^{B})\cdot 2^{L_\ell}\bigr)$ & $p\in[1,2^{B}{-}2]$                 & $L_\ell{=}5$ & $[256]$   & ${\sim}256$\,B \\
\bottomrule
\end{tabular}
\end{table}

\paragraph{Softmax forward.}
\label{app:poly_softmax}
Four integer steps per row:
\begin{enumerate}
    \item $x'_i=x_i-\max_j x_j\in[-255,0]$.
    \item $E_i=\textsc{LUT}_{\exp}[e_x,x'_i]$.
    \item $Z=\sum_j E_j$ and $\mathrm{inv}_Z=\lfloor 2^{S_R}/Z\rfloor$ with $S_R{=}30$ (one INT32 divide per row; the $K$ normalizations below are multiplies).
    \item $p_i=\textsc{shift\_round}(E_i\cdot\mathrm{inv}_Z,\;S_R-B)$, yielding unsigned int$B$ in $[0,2^{B}{-}2]$ (the $2^{L}$ scale of $E_i$ and $Z$ cancels in the ratio; stochastic rounding is applied during training).
\end{enumerate}
The same cached int$B$ tensor $p$ is reused by the backward and the loss below.

\paragraph{Loss value (reporting only).}
\label{app:ce}
For logging only. For each sample we index $p_{n,y_n}\in[1,2^{B}{-}2]$ (floored to $\ge 1$ so $-\log 0$ never appears) into $\textsc{LUT}_{\log}$ and take the integer batch mean
\begin{equation}
\mathcal{L} \;=\; \tfrac{1}{N}\sum_{n=1}^{N}\textsc{LUT}_{\log}[p_{n,y_n}],\qquad e_{\mathcal{L}}=-L_\ell,
\end{equation}
with division by $N$ via \Cref{eq:div_pow2} or \Cref{eq:div_reciprocal}.

\paragraph{Gradient (seed of backpropagation).}
At the loss head, softmax and cross-entropy telescope into
\begin{equation}
\bigl(\partial\mathcal{L}/\partial X_L\bigr)_{n,k} \;=\; \tfrac{1}{N}\bigl(p_{n,k}-\mathbf{1}[k{=}y_n]\bigr),
\end{equation}
a signed integer subtract on the int$B$ tensor $p$ (promoted to int$(B{+}1)$ to carry the $-1$) followed by division by $N$.

Away from the loss (softmax inside attention), we apply the Jacobian on the cached $p$: $a=\sum_j p_j\,\partial y_j$ (product + sum), then $\partial x_i=p_i(\partial y_i-a)$ (subtract + product), with one compaction back to int$B$.

\subsection{Vision Transformer block}
\label{app:vit}

The ViT reuses the same primitives as the CNN models: patch embedding is a strided convolution (integer product), and the CLS token and positional embeddings are learned integer parameters merged with the sum primitive. Each pre-norm block is
\begin{equation}
Y = X + \textsc{MHA}(\textsc{LayerNorm}(X)),
\qquad
Z = Y + \textsc{MLP}(\textsc{LayerNorm}(Y)),
\end{equation}
with LayerNorm from \Cref{app:layernorm}, integer GELU inside the MLP from \Cref{app:activations}, and the residual adds folded into the sum primitive (\Cref{eq:sum_align}).

\paragraph{Multi-head attention, forward.}
Per head (hidden size $d_h$, sequence length $T$):
\begin{enumerate}
    \item Three integer linear projections on the layer-normed input produce $(Q,e_Q)$, $(K,e_K)$, $(V,e_V)$ in $\mathbb{Z}^{T\times d_h}$.
    \item Score accumulator $A_{QK}=QK^{\top}\in\mathbb{Z}^{T\times T}$ (INT32 product, \Cref{eq:prod_acc}), with natural exponent $e_Q+e_K$.
    \item Scale by $1/\sqrt{d_h}$ and compact to int$B$:
    \begin{equation}
        (C_{\mathrm{score}},\,e_{\mathrm{score}})=\textsc{compact}\!\bigl(\textsc{shift\_round}(A_{QK}\cdot r_d,\,s_d),\; e_Q+e_K\bigr),
    \end{equation}
    where $(r_d,s_d)$ encode $1/\sqrt{d_h}$ as a rational shift (see next paragraph).
    \item Row-wise integer softmax (\Cref{app:poly_softmax}) yields $p\in[0,2^{B_p}{-}1]^{T\times T}$.
    \item Context accumulator $A_{\mathrm{ctx}}=pV\in\mathbb{Z}^{T\times d_h}$, compacted to int$B$ with natural exponent $-B_p+e_V$.
    \item A final integer linear concatenates heads and projects back to the embedding dimension.
\end{enumerate}
Every intermediate is an $(C,e)$ pair; only $p$ and $V$ need to be cached for the backward.

\paragraph{$1/\sqrt{d_h}$ as a rational shift.}
Since $d_h$ is fixed at model-design time (commonly $d_h\in\{32,64,96\}$), $1/\sqrt{d_h}$ is a compile-time constant. We represent it as $r_d/2^{s_d}$ with $r_d=\lfloor 2^{s_d}/\sqrt{d_h}\rfloor$: for ViT-B ($d_h{=}64$), $s_d{=}10$ and $r_d{=}128$ give $128/1024=1/\sqrt{64}$ exactly; at the same $s_d$, $d_h{=}32$ and $d_h{=}96$ are lossless within one LSB. The backward applies the identical $(r_d,s_d)$ to the score gradient.

\paragraph{Multi-head attention, backward.}
Given the upstream gradient $\partial Z$ and the cached $(p, V)$, the reverse pass walks the forward pipeline in reverse, each stage using only product, sum, and shift-round primitives:
\begin{enumerate}
    \item Output projection. Back through the final linear yields the context gradient $\partial A_{\mathrm{ctx}}\in\mathbb{Z}^{T\times d_h}$ (one integer product) and the projection weight gradient.
    \item Context matmul. The product $A_{\mathrm{ctx}}=pV$ splits into $\partial p=\partial A_{\mathrm{ctx}}\,V^{\top}$ and $\partial V=p^{\top}\partial A_{\mathrm{ctx}}$.
    \item Softmax Jacobian. Row-by-row on the cached $p$: $\partial C_{\mathrm{score},i}=p_i\bigl(\partial p_i-\sum_j p_j\,\partial p_j\bigr)$, exactly as in \Cref{app:output_head}.
    \item Score scaling. The same rational shift is applied symmetrically: $\partial A_{QK}=\textsc{shift\_round}(\partial C_{\mathrm{score}}\cdot r_d,\,s_d)$.
    \item Score matmul. $A_{QK}=QK^{\top}$ splits into $\partial Q=\partial A_{QK}\,K$ and $\partial K=\partial A_{QK}^{\top}Q$.
    \item Q/K/V projections. Back through the three input projections produces input-gradient streams $\partial X^{(Q)},\partial X^{(K)},\partial X^{(V)}$ and weight gradients $\partial W_Q,\partial W_K,\partial W_V$.
    \item Stream merge. Two applications of the sum primitive (\Cref{eq:sum_align}) give $\partial X=\partial X^{(Q)}+\partial X^{(K)}+\partial X^{(V)}$.
\end{enumerate}
Each INT32 accumulator is compacted back to int$B$ via \Cref{eq:compact_tensor} before storage, so every tensor persisted across stages is an $(C, e)$ pair.


\section{Training memory computation}
\label{app:memory}

The \textbf{Training memory (GB)} column of \Cref{tab:main_results} reports the total training memory of each method at the benchmark configuration of its row. This appendix lists the per-tensor storage format of each method and the formula we use to combine it with the model's parameter count and activation footprint.

\paragraph{Per-tensor storage format.}
The per-tensor formats assumed for each method are listed in \Cref{tab:storage_formats} (main text). Summing the weight, gradient, and velocity widths gives the persistent state in bytes per parameter; the activation format scales with batch size and input resolution. Two points deserve comment.
FP8 NVIDIA~\cite{micikevicius2022fp8formatsdeeplearning} quantizes to FP8 only at GEMM inputs; outputs and saved-for-backward activations remain FP16/BF16 (their \S4.1 states ``output tensors were left in higher precision\ldots moving more tensors to FP8 is the subject of future study'').
INT8 forward/backward methods~\cite{zhu2019unifiedint8trainingconvolutional,zhao2021distributionadaptiveint8quantization,jin2022f8netfixedpoint8bitmultiplication} and Ghaffari~\cite{ghaffari2023integerarithmeticdeeplearning} are implemented as GPU emulations: the reduced-precision format is used at GEMM operand boundaries, while the tensor carried between layers (and saved for backward) is kept at FP32 in the implementation that produced their reported accuracy.
The 8-bit claims in those papers refer to the arithmetic datapath, not to tensor storage.

\paragraph{Computing total training memory.}
For a model with $P$ parameters and a forward pass that saves $A$ activation elements per sample (summed across layers), training at batch size $N$ uses
\begin{equation}
M_\text{train} \;=\; b_\text{param}\cdot P \;+\; b_\text{act}\cdot A\cdot N,
\label{eq:mem_formula}
\end{equation}
where $b_\text{param}$ is the persistent state in bytes per parameter from \Cref{tab:storage_formats} and $b_\text{act}$ is the activation bytes per element from the same table. For the FP32 baseline, $b_\text{param}=12$ and $b_\text{act}=4$.

\paragraph{Per-configuration values.}
\Cref{tab:mem_inputs} lists $P$, $A$, and $N$ for every benchmark configuration used in \Cref{tab:main_results}. The values of $P$ are standard for the corresponding reference implementations; the activation counts $A$ are obtained from PyTorch autograd profiling on the respective architectures, counting only the tensors that autograd retains for the backward pass.

\begin{table}[h]
\centering
\caption{Per-configuration inputs to \Cref{eq:mem_formula}: parameters $P$, saved-for-backward activations $A$ per sample, batch size $N$.}
\label{tab:mem_inputs}
\small
\begin{tabular}{lccc}
\toprule
\textbf{Configuration} & \textbf{$P$ (M)} & \textbf{$A$ (M/sample)} & \textbf{$N$} \\
\midrule
ResNet-18 / CIFAR-10 (32$^2$)            & 11.2 & 1.29 & 256 \\
MobileNetV2 / CIFAR-10 (32$^2$)          & 2.24 & 1.88 & 256 \\
ViT-Ti / CIFAR-10 (32$^2$)               & 5.36 & 7.91 & 256 \\
ViT-B / CIFAR-10 fine-tune (224$^2$)     & 86.0 & 35.1 & 64  \\
ResNet-18 / CIFAR-100 (32$^2$)           & 11.2 & 1.29 & 256 \\
ResNet-34 / CIFAR-100 (32$^2$)           & 21.3 & 2.01 & 256 \\
ResNet-18 / ImageNet (224$^2$)           & 11.7 &  4.7 & 256 \\
ResNet-50 / ImageNet (224$^2$)           & 25.6 & 21.5 & 256 \\
ResNet-101 / ImageNet (224$^2$)          & 44.5 & 31.8 & 256 \\
\bottomrule
\end{tabular}
\end{table}

Plugging each (method, configuration) pair into \Cref{eq:mem_formula} with the bytes from \Cref{tab:storage_formats} reproduces the \textbf{Training memory (GB)} entries of \Cref{tab:main_results}. The computation is implemented as a short Python script released with the paper.

\section{Training recipe and hyperparameters}
\label{app:recipe}
This appendix collects the training details omitted from the main text: how inputs and activations are quantized, how biases and weight decay are folded into the integer pipeline, the integer learning-rate schedule, and the full hyperparameter grid of \Cref{tab:hparams}.

\paragraph{Hardware and software.}
All training runs use a single NVIDIA A100-SXM4 (40\,GB) GPU. The integer pipeline is implemented on top of PyTorch~2.10 with CUDA~12.8 and cuDNN~9.1.

\paragraph{Input quantization.}
The dataloader produces float RGB tensors in $[0,1]$; we normalize with standard per-channel statistics and then quantize once to INT8 at the framework boundary using the compaction primitive (\Cref{eq:compact_tensor}):
\begin{equation}
(X_0,\,e_{X_0}) \;=\; \textsc{compact}\!\bigl(\lfloor x\cdot 2^{7}\rfloor,\; -7\bigr).
\end{equation}
The shift is data-independent because the normalized range is bounded, so no online calibration is required. This is the only float-to-integer conversion in the pipeline; at evaluation the INT8 logits $X_L$ feed integer $\arg\max$ directly (\Cref{app:ce}), so no integer-to-float conversion is needed either.

\paragraph{Activation quantization.}
Between layers, activations are re-quantized via the compaction primitive (\Cref{eq:compact_tensor}); the shift $S$ is computed from the per-tensor $\|A\|_\infty$ during the forward pass, so no separate calibration step is needed. For ViT-B we additionally clip activations at the $99.99^{\text{th}}$ percentile of absolute values (\texttt{scale\_percentile=0.9999}) before computing the shift, preventing a single rare outlier from consuming the dynamic range. CNNs and ViT-Ti do not require this clipping at any of the bit widths we tested.

\paragraph{Biases (ViT only).}
The CNN models in our setup (ResNet-18, MobileNetV2) omit convolution and linear biases because batch normalization absorbs them; ViT-B retains biases in its Q/K/V and MLP projections. Each bias is stored as a $(b, e_b)$ pair at the weight bit width (INT8 by default) and aligned to the output exponent $e_{XW}$ via the bounded-shift policy of \Cref{sec:op_sum} before being added to the pre-activation.

\paragraph{Momentum and weight decay.}
The momentum coefficient $\beta$ is stored as the integer pair $(\beta_{\text{num}}, \beta_{\text{shift}})$ with $\beta \approx \beta_{\text{num}}\,2^{-\beta_{\text{shift}}}$; the default $\beta{=}0.9$ uses $(\beta_{\text{num}}, \beta_{\text{shift}}){=}(230, 8)$ giving $\beta\approx 230/2^8$. $L_2$ regularization is injected into the per-group gradient \emph{before} the momentum update, so the decay term participates in the velocity accumulation:
\begin{equation}
G_k \;\leftarrow\; G_k \,+\, \textsc{shift\_round}\!\bigl(\lambda_{\text{num}}\cdot W_k,\; \lambda_{\text{shift}}\bigr),
\end{equation}
with $(\lambda_{\text{num}}, \lambda_{\text{shift}})$ precomputed for the target $\lambda$ (e.g.\ $\lambda{=}10^{-4}$). The augmented $G_k$ then feeds \Cref{eq:vel_compact}, so no separate weight-decay step is applied at the weight update itself.

\paragraph{Integer learning-rate schedule.}
The learning rate is stored as $\eta\approx\eta_{\text{num}}\,2^{-\eta_{\text{shift}}}$. Since schedules are known before training, we precompute one $(\eta_{\text{num}}^{(t)}, \eta_{\text{shift}}^{(t)})$ pair per iteration $t$ as the best-fit
\begin{equation}
(\eta_{\text{num}}^{(t)},\,\eta_{\text{shift}}^{(t)}) \;=\; \arg\min_{\eta_{\text{num}},\,\eta_{\text{shift}}}\,\bigl|\eta_{\text{num}}\,2^{-\eta_{\text{shift}}}-\eta^{(t)}\bigr|
\quad\text{subject to}\quad 1\le\eta_{\text{num}}\le\eta_{\text{num},\max},
\end{equation}
where $\eta^{(t)}$ follows the regime in use: geometric decay $\gamma\eta$ at milestones for step schedules; $\eta_{\min}+\tfrac{1}{2}(\eta_{\max}-\eta_{\min})(1+\cos(\pi t/T))$ for cosine; linear interpolation during warmup.

\begin{table}[h]
\centering
\caption{Per-model hyperparameters used for \Cref{tab:main_results}. All runs: SGD with momentum, cosine LR with linear warmup.}
\label{tab:hparams}
\small
\begin{tabular}{lccccc}
\toprule
& \textbf{RN-18 C10/C100} & \textbf{MNv2 C10} & \textbf{ViT-Ti C10} & \textbf{ViT-B C10} & \textbf{RN-18 INet} \\
\midrule
Batch size           & 256 & 256 & 256 & 64  & 256 \\
Epochs               & 200 & 200 & 400 & 50  & 90 \\
Base LR              & 0.1 & 0.05 & 0.05 & 0.001 & 0.1 \\
Warmup epochs        & 5   & 5   & 5   & 0   & 5 \\
Momentum $\beta$     & 0.9 & 0.9 & 0.9 & 0.9 & 0.9 \\
Weight decay         & $10^{-4}$ & $10^{-4}$ & $10^{-4}$ & $10^{-4}$ & $10^{-4}$ \\
Velocity bits $B_V$  & 8   & 8   & 8   & 8   & 8 \\
Accumulator bits     & 32  & 32  & 32  & 32  & 32 \\
Activation clip pct. & --- & --- & --- & 0.9999 & --- \\
\multirow{2}{*}{Augmentation}         & \multirow{2}{*}{crop+flip} & crop+flip & crop+flip & \multirow{2}{*}{RandAugment} & crop+flip \\
       & & +cutout & +AutoAug &  & +AutoAug \\
\bottomrule
\end{tabular}
\end{table}

\section{Additional experiments}
\label{app:ablations}

\begin{figure}[h]
\centering
\includegraphics[width=0.95\textwidth]{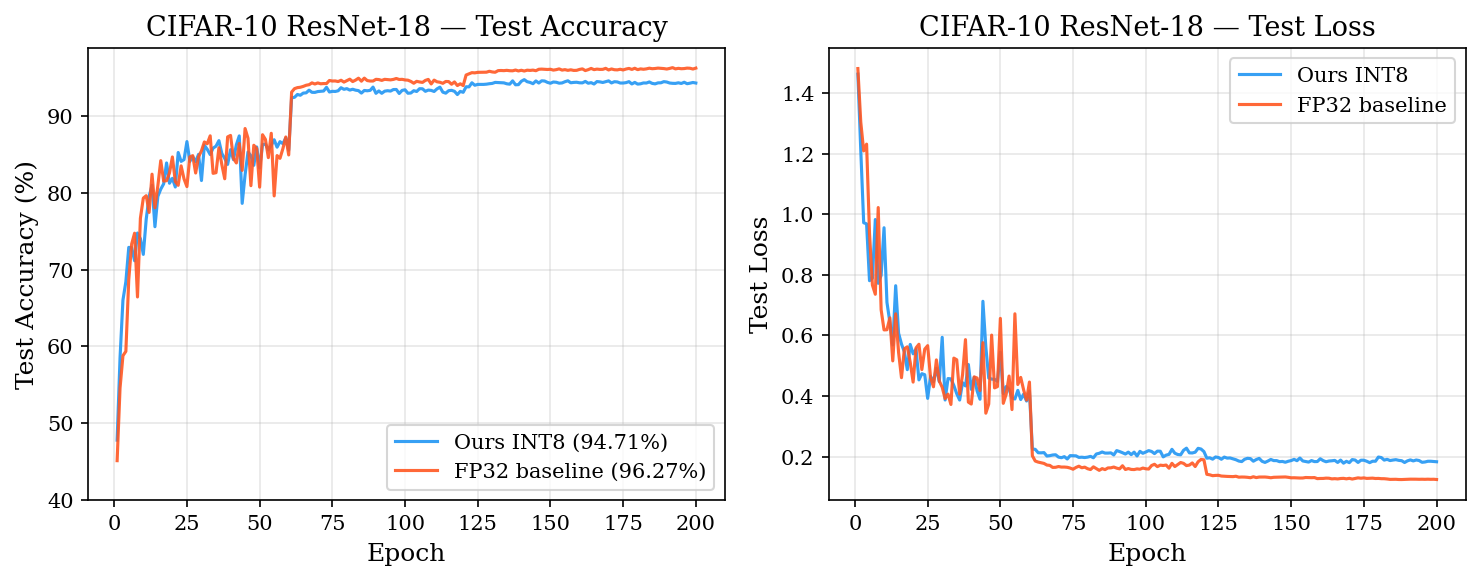}
\caption{ResNet-18 / CIFAR-10: integer-only (INT8) vs.\ FP32 baseline, 200 epochs, step LR at 60/120/180.}
\label{fig:training_curves}
\end{figure}

This appendix supplements \Cref{sec:results} with: (i) training dynamics for ResNet-18/CIFAR-10 (\Cref{fig:training_curves}); (ii) an operator-level breakdown of the ViT-B accuracy gap on CIFAR-10 (\Cref{tab:vit_ablation}); and (iii) a layer-by-layer cosine-similarity analysis of the INT8 ViT that motivates the need for fine-tuning.

\paragraph{Training dynamics.}
\Cref{fig:training_curves} shows that INT8 training tracks the FP32 baseline on ResNet-18/CIFAR-10 over 200 epochs; step-LR drops produce clean discrete transitions with no instability or divergence.

\paragraph{Where the INT8 ViT drifts.}
Table~\ref{tab:vit_ablation} attributes the entire 3.2\% pre-finetuning gap to the Linear projections: replacing only the Linear layers with INT8 (keeping LayerNorm, GELU, and softmax in float) costs 3.2\%, while subsequently moving LayerNorm, GELU, and softmax into the integer pipeline costs nothing further (LayerNorm $-0.1$, GELU $+0.1$, softmax $+0.2$). In other words, the integer attention machinery introduced in this paper -- LUT-based softmax, $1/\sqrt{d_h}$ as a rational shift, and the cached-$p$ Jacobian backward -- is essentially free in accuracy terms; the residual gap is the generic Linear-quantization error that any INT8 fwd/bwd method incurs.

\begin{table}[h]
\centering
\caption{Operator layer ablation on ViT-B fine-tuning (CIFAR-10). All rows use INT8 weights and activations.}
\label{tab:vit_ablation}
\begin{tabular}{cccccc}
\toprule
\textbf{Softmax} & \textbf{GELU} & \textbf{LayerNorm} & \textbf{Linear} & \textbf{Acc.\ (\%)} & \textbf{$\Delta$ vs.\ float} \\
\midrule
float & float & float & float  & 99.1 & (float baseline) \\
float & float & float & INT8   & 95.9 & $-3.2$ \\
float & float & INT8  & INT8   & 95.8 & $-3.3$ \\
float & INT8  & INT8  & INT8   & 95.9 & $-3.2$ \\
INT8  & INT8  & INT8  & INT8   & 96.1 & $-3.0$ \\
\midrule
\multicolumn{4}{l}{\cite{ghaffari2023integerarithmeticdeeplearning} (float softmax, FP32 weights)} & 98.8 & $-0.3$ \\
\bottomrule
\end{tabular}
\end{table}

A complementary cosine-similarity analysis of the INT8 ViT without fine-tuning shows where this Linear-quantization error compounds. The match to the float model degrades through the stack from $0.977$ after the patch embedding to $0.83$ after block~0, $0.64$ after block~2, and below $0.01$ after block~5. The drop is concentrated inside the attention sublayer -- $0.95 \to 0.64$ in a single block -- but, consistent with Table~\ref{tab:vit_ablation}, the source is not the attention-specific operators: it is the four Linear projections per block ($QW_Q$, $KW_K$, $VW_V$, and the output projection), whose ${\sim}72$ INT8 compaction points compound multiplicatively across the 12-block stack. Fine-tuning then closes the gap as the weights adapt to the INT8 quantizer.


\newpage

\end{document}